\documentclass{ifacconf}

\usepackage{graphicx}      
\usepackage{natbib}        
\usepackage{makecell}
\usepackage{enumitem}
\usepackage{algorithm, algorithmic}
\usepackage{amsmath,amssymb,amsfonts}
\usepackage{multirow}
\usepackage{xurl}

\newcommand{\bianca}{\renewcommand\NAT@open{[}\renewcommand\NAT@close{]}}

\begin{document}
\begin{frontmatter}

\title{Learning from Demonstrations of Critical Driving Behaviours Using Driver's Risk Field\thanksref{footnoteinfo}} 

\thanks[footnoteinfo]{This work was carried out within the thesis of Yurui Du at Siemens as part of the FOCETA project. This project was funded by the European Union’s Horizon 2020 research and innovation programme under grant agreement No 956123. The work is also benefited from the Flanders Innovation \& Entrepreneurship – VLAIO funded project BECAREFUL. © 2023 the authors. This work has been accepted to IFAC for publication under a Creative Commons Licence CC-BY-NC-ND}

\author[aff01,aff02]{Yurui Du} 
\author[aff02]{Flavia Sofia Acerbo} 
\author[aff01]{Jens Kober}
\author[aff02]{Tong Duy Son}

\address[aff01]{Department of Cognitive Robotics, Delft University of Technology, Delft, the Netherlands; E-mail: duyurui10@gmail.com; j.kober@tudelft.nl}
\address[aff02]{Siemens Digital Industries Software, Leuven, Belgium, E-mail: flavia.acerbo@siemens.com; son.tong@siemens.com}

\begin{abstract}                
In recent years, imitation learning (IL) has been widely used in industry as the core of autonomous vehicle (AV) planning modules. However, previous IL works show sample inefficiency and low generalisation in safety-critical scenarios, on which they are rarely tested. As a result, IL planners can reach a performance plateau where adding more training data ceases to improve the learnt policy. First, our work presents an IL model using the spline coefficient parameterisation and offline expert queries to enhance safety and training efficiency. Then, we expose the weakness of the learnt IL policy by synthetically generating critical scenarios through optimisation of parameters of the driver's risk field (DRF), a parametric human driving behaviour model implemented in a multi-agent traffic simulator based on the Lyft Prediction Dataset. To continuously improve the learnt policy, we retrain the IL model with augmented data. Thanks to the expressivity and interpretability of the DRF, the desired driving behaviours can be encoded and aggregated to the original training data. Our work constitutes a full development cycle that can efficiently and continuously improve the learnt IL policies in closed-loop. Finally, we show that our IL planner developed with less training resource still has superior performance compared to the previous state-of-the-art.
\end{abstract}


\end{frontmatter}

\section{Introduction}



Today, autonomous vehicles (AVs) worldwide are undergoing extensive road tests in the real world, and some have already been put into active service. However, level 4+ autonomous driving still remains a significant challenge due to the ``long tail" of real-world driving events, meaning AVs can be unsafe in rarely occurring safety-critical scenarios \citep{Jain2021}. In the AV application stack, the motion planning module is one of the keys to solving this bottleneck as it determines the AV's driving policy. By learning from large-scale driving datasets of expert demonstrations, imitation learning (IL) has been exploited as the core planner in real-world traffic scenarios, such as unsigned rural roads \citep{alvinn1989}, highways \citep{nvidia2016}, and urban driving \citep{wayve2020,Waymo2019,Lyft2022}. 


\begin{figure*}[!htbp]
  \centering
  \includegraphics[width=0.8\textwidth]{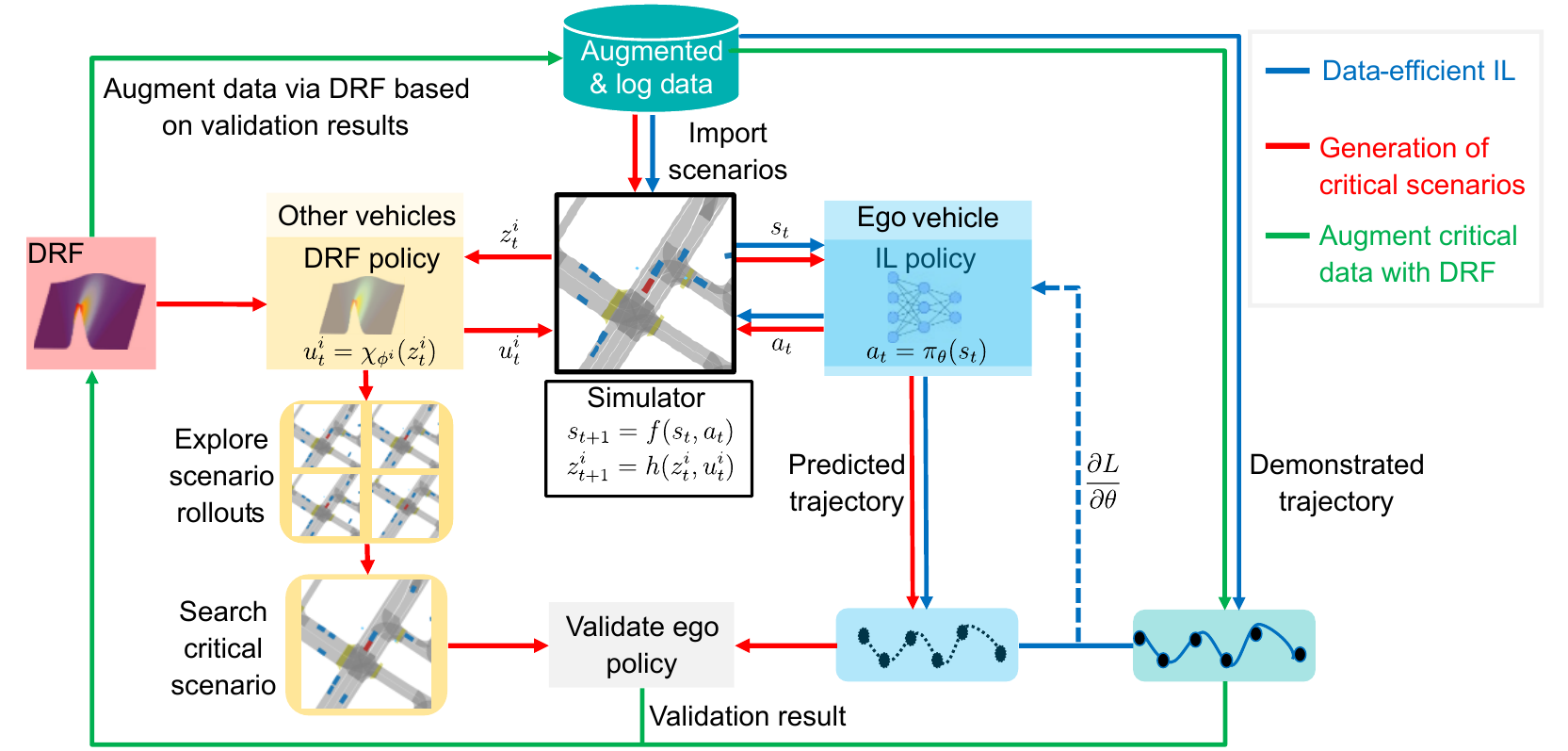}
  \caption{Overview of our three-part work that respectively addresses a safe, efficient IL method, generation of critical scenarios for validation, and data augmentation encoding desired driving behaviours via DRF. All three parts constitute a development cycle that allows us to continuously improve IL policies in closed-loop.}
  \label{overview}
\end{figure*}

However, despite the growing use of IL in AVs' planning module, efficiently improving its safety in long-tail events is a difficult task. In our view, it can be decomposed into three sub-tasks, namely, data-efficient learning, generation of safety-critical scenarios, and data augmentation of critical driving behaviours. Our motivations are given as follows:

It is observed that IL models require an excessive amount of training resource in order to achieve capable, but sometimes unsafe driving behaviours due to the distributional shift between the training and validation distributions \citep{Waymo2019,Lyft2022}. To enhance training efficiency and driving safety, we utilise the spline parameterisation for the IL model's predicted trajectory as proposed by our previous work \citep{spl2020} and an offline expert query approach to mitigate the distributional shift \citep{Lyft2022}.

Validation of IL models under critical traffic scenarios is often missing in published research. Most IL models are validated with log-replay data, where the traffic agents' trajectories are logged, and the dynamic interactions between traffic agents are not considered. To address this problem, recent research proposed to build reactive simulations with traffic agents that respond to others \citep{simnet2021,trafficsim2021,scenegen2021,symphony2022}. These works however mainly focused on generating traffic scenarios similar to the ones from the original dataset, which are not representative of critical traffic scenarios. Critical scenarios can also be manually designed by human experts, but this approach is not scalable to generate highly complex urban traffic scenarios with multiple traffic agents \citep{abey2019}. Furthermore, the diverse driving styles of traffic agents in the real world are not considered, making the obtained critical scenarios not satisfactorily capture the diversity and complexity of real-world driving. To generate realistic, complex critical scenarios that can help us discover weak driving policies in validation, we employ driver's risk field (DRF) \citep{DRF2020}, a parametric model that represents human driving behaviours using the driver's subjective perceived risk of the environment. Compared to other driver models discussed in Sec. \ref{driver models comparison}, DRF is a unified theory that allows us to represent different driving behaviours by tuning few parameters without switching between different models. We use DRF as traffic agents in a model-based multi-agent simulator based on the Lyft Prediction Dataset. By optimising the DRF parameters, critical traffic scenarios can be generated with realistic and diverse agents on a large scale. 

Another bottleneck for IL is that increasing size of dataset does not necessarily improves IL models' robustness and safety \citep{Lyft2022}. This may indicate that IL models reach a performance plateau during training and stop learning from normal traffic data. To continuously improve the performance of IL models, we present a novel and flexible data augmentation method in which the DRF is exploited to encode desired driving behaviours in the original training data to purposefully improve poorly trained IL policies exposed in the validation results from critical scenarios. 


Our contributions are three-fold, as summarised in Fig. \ref{overview}:

\begin{enumerate}
\item{An IL method combining the spline coefficient parameterisation with the closed-loop offline expert query approach for efficient training. We demonstrate its superior performance over existing methods by validating it in large urban driving datasets and our generated critical traffic scenarios.}
\item{Scalable generation of realistic and critical traffic scenarios in an interactive DRF-based traffic simulator to test ego driving policies in validation.}
\item{A novel data augmentation method encoding desired driving behaviours with DRF to continuously improve IL performance in both recorded scenarios from logged data and our generated critical scenarios.}
\end{enumerate}

\section{Related Works}

In this section, we first discuss imitation learning and useful approaches to improve performance. Second, limitations of previous works on critical scenario generation are presented. Third, our motivations for modelling traffic agents using DRF are given via a comparison of different human driving behaviour models.

\subsection{Imitation learning}

Compared to optimisation-based motion planners, IL is attractive for its scalability to integrate new functionalities by learning from expert demonstrations rather than optimising human engineered objective functions. Additionally, with the availability of large-scale driving datasets and continuously produced data during testing, IL is becoming a popular method for motion planning in AV industry. However, one of the major challenges is that IL often suffers from distributional shift, which is caused by the compounding error in the sequential decision-making process, such as motion planning for AVs. It leads the ego vehicle to unfamiliar scenarios that are not included in the training distribution. Eventually, the behaviour of the ego vehicle becomes completely unpredictable and unsafe due to large deviations from the demonstration. 

In practice, many approaches have been proposed to mitigate the distributional shift and significantly improve IL performance. While these approaches may seem different, they mostly mitigate the distributional shift by providing corrective actions during training so that the ego vehicle learns to recover from earlier deviations in the sequential decision-making process. One approach \citep{Waymo2019} leverages simple behaviour cloning with data augmentation by adding perturbation noise to provide more robust driving policies. Similarly, another approach \citep{nvidia2016} tries to directly label the perturbed camera images with corrective actions to avoid drifting. However, these approaches generally depend on empirical experiences to engineer noise mechanisms before training. A more theoretically satisfying approach is the dataset aggregation (DAgger) \citep{DAgger2011}, \citep{Lyft2022}, which generates the training distribution of corrective actions on the run and guarantees an ideal linear regret bound to mitigate the distributional shift. However, it also exerts a heavier computation burden. To improve training efficiency, spline parameterisation as a powerful representation of predicted trajectories for IL has been proposed \citep{spl2020}. 






\subsection{Critical scenario generation}

Prior to our work, critical scenario generation has been studied in \citep{abey2019}. By manually assigning waypoints and multiple available actions for each agent to choose from during the rollout of their policies, the most critical scenario can be found using search algorithms. However, the excessive manual labour required and the heavy computation burden greatly impair its scalability because the time complexity of this approach grows exponentially with the number of traffic agents, designed waypoints and actions. Other works on traffic scenario generation and reactive simulation with interactive agents mostly used generative methods such as latent variable models \citep{trafficsim2021}, autoregressive models \citep{scenegen2021}, and generative adversarial imitation learning \citep{simnet2021}, in order to capture the possibility of multiple futures. However, these works mainly focused on generating similar traffic scenarios or agents with similar driving policies as demonstrated in the original dataset. Therefore, the generated scenarios are not critical scenarios that are purposefully designed to challenge weak driving policies in validation.



\subsection{Realistic traffic agent modelling}
\label{driver models comparison}

Human driving behaviour models can be categorised into non-parametric and parametric ones \citep{Lefevre2014}. For non-parametric approaches, the driver model relies on a large amount of data to learn policies that behave like human driving. Whereas for the parametric models, the driver behaviour model is often built from prior expert knowledge to capture human driving features in mathematically analytical forms, in which the parameters can be identified by fitting the model to the given data. 

The decision-making process of non-parametric models is often considered a blackbox, meaning their driving behaviours cannot be easily adjusted with guaranteed legality. Moreover, as diverse behaviours for different traffic agents need to be modelled individually to make realistic simulations, it will be extremely expensive to train non-parametric models for all driving behaviours.


Compared to non-parametric models, building parametric human driving behaviour models requires considerably smaller amount of parameters, and these parameters often have clear physical and mathematical meanings, making it easier to interpret and control behaviours of the model. However, most parametric models only work for specific driving scenarios, such as car following in free roads \citep{treiber2000} or multi-lane highways \citep{kesting2007}. These fragmented methods are by nature flawed because real driving scenarios are highly complex. Therefore, it is difficult to identify all possible traffic scenarios and design smooth transitions for them.


For these reasons, the driver's risk field (DRF) \citep{DRF2020}, a parametric human driving behaviour model, is especially suitable for realistic agent modelling because: 

\begin{enumerate}
\item{It provides the driver’s subjective view of the driving risk in any given scenario.} 
\item{It can explain diverse driving behaviours with a unified theory.}
\item{It has interpretable parameters tunable to mimic diverse driving behaviours.}
\end{enumerate}










\section{Methods}
In this section, we first specify the formulation of our IL method for the ego vehicle. Then, the parametric modelling of other traffic agents using DRF is discussed. We propose our method to generate critical scenarios with DRF agents that act adversarially to challenge the IL policy in validation. Finally, we present a novel data augmentation method that encodes demonstrations of critical driving behaviours to purposefully improve weak IL policies exposed in critical scenarios.

\subsection{Efficient IL with spline coefficient parameterisation and closed-loop offline expert query}
IL is a supervised learning method that aims to directly mimic driving behaviours from expert demonstrations. In the context of IL, the expert policy is defined as $a_{t}^{*}=\pi^{*}(s_{t})$, i.e., the mapping from an agent's states to its actions. We adopt a similar approach to the one proposed in \citep{Lyft2022}, which is similar to DAgger \citep{DAgger2011}, but with better computational efficiency owing to an offline synthetic expert query rather than an active expert policy to aggregate training datasets. This offline expert query approach is achieved by a closed-loop training scheme. Assuming that the dataset $D^*$ consists of $N$ expert trajectories and each trajectory $\tau_{i}$ has the length of $T$ steps, namely $D^{*} = \{\tau_{i}\}_{i=1}^{N}$, $\tau_{i} = \{(s_{i,j}, a_{i,j})\}_{j=1}^{T}$, we first sample the ego vehicle's current policy for K steps, which will lead the ego vehicle to unfamiliar scenarios due to the distributional shift. Then, the current policy is updated by minimising the above loss function in the remaining $T-K$ steps so the ego vehicle learns to recover from mistakes caused by the distributional shift. The parameter of the policy network is denoted by $\theta$, which can be learnt by minimising the discounted cumulative expected loss with a discount factor of $\gamma$:
\begin{equation}
    \hat{\theta}=\arg \min_{\theta} \mathbb{E}_{\tau\sim\pi^{*}} \sum_{t=K}^{T}\gamma^{t-K}L (\pi_{\theta}(s_{t}), a_{t}^{*})\,.
\label{objective function}
\end{equation}

In this work, $L (\pi_{\theta}(s_{t}), a_{t}^{*}) = ||\pi_{\theta}(s_{t}) - a_{t}^{*}||_{1}$ is the L1 distance between the learner's action $\pi_{\theta}(s_{t})$ and the expert action $a_{t}^{*}$. Similarly to \citep{spl2020}, spline coefficients are used to parameterise trajectories in the dataset $D^*$ instead of using discrete waypoints for better safety, more stable long-horizon predictions and smoother trajectories. Furthermore, we show that this parameterisation greatly improves the training efficiency in Sec. \ref{training efficiency}. 

The states $s_t$ and actions $a_t$ of the original expert trajectories are both denoted as a 3D vector $(x, y, \alpha)$ representing the position and orientation in the $SE(2)$ space. The corresponding $n$ spline coefficients in all three directions can be expressed as a matrix $\mathrm{A}_{3\times n}$. By replacing $a_{t}^{*}$ in \eqref{objective function} with $\mathrm{A}_{3\times n}^{*}$, our objective function can be rewritten as:
\begin{equation}
    \hat{\theta}=\arg \min_{\theta} \mathbb{E}_{\tau\sim\pi^{*}} \sum_{t=K}^{T}\gamma^{t-K}L (\pi_{\theta}(s_{t}), \mathrm{A}_{3\times n}^{*})\,.
\label{objective function A}
\end{equation} 



\subsection{Parametric agent modelling using DRF}
\label{drfmodelling}
The DRF builds the driver's subjective view of its surrounding environment as a 2D Gaussian distribution along the predicted path. The perceived risk is derived from DRF representing the driver's subjective view of the driving risk in traffic. It is a function of the ego vehicle's current velocity and steering angle $P_{risk}(v, \delta)$. Then, based on the risk threshold theory, the future velocity and steering angle are obtained by solving an optimisation problem to keep the perceived risk below the assigned threshold. For detailed formulations of the DRF, please refer to \citep{DRF2020}. 


\subsection{Critical scenario generation}
In this part, we detail how to generate critical traffic scenarios with agents that follow DRF policies controlling their velocity profiles along their original trajectories. The agents are designed to react adversarially to the ego vehicle's driving policy by optimising the DRF parameters of agents. The traffic scenarios are initialised based on real-world urban driving data to improve the complexity and realism of the generated scenarios.

We assume $s_t = \{x_t, y_t, \alpha_t\}$ to be the state vector of the ego vehicle's pose at time t. This vector includes the 2D position and orientation of the vehicle w.r.t. the ego-centric reference frame at $t=0$. Let $Z_t = \{z_t^i\}_{i=1}^{M}$ be the state vector that consists of the pose of all other $M$ agent vehicles closest to the ego vehicle, where $z_{t}^{i}$ is the state vector of the $i^{th}$ agent vehicle's pose. Let us assume that $a_{t} = \pi_{\theta}(s_t)$ is the IL policy of the ego vehicle and $u_{t}^{i} = \chi_{\phi^{i}}(z_t^i)$ is the parametric policy of the $i^{th}$ agent parameterised by $\phi^{i}$, and also the dynamics model of the ego vehicle $s_{t+1} = f(s_t, a_t)$ and the agent $z_{t+1}^{i} = h(z_t^i, u_t^i)$. We can obtain $\phi^{i}$ for each agent's parametric policy leading to critical traffic scenarios by optimising the following objective:
\begin{equation}
    \Phi^{*} = \arg \min_{\Phi} J(\theta, \Phi)\,,
\label{critical objective}
\end{equation} 
where $\Phi = \{\phi^{i}\}_{i=1}^{M}$ is the vector of agents' parameters and $J(\theta, \Phi)$ is the cost-to-go function computed from the scenario via unrolling all vehicles' policies. The cost is computed from the L1 distance between the ego vehicle and other vehicles and the total number of accidents (collisions, off-road incidents) to encourage the formation of dense traffic and collisions:
\begin{equation}
    J(\theta, \Phi) = \mathbb{E}_{s_t, Z_t}\sum_{t=0}^{T} L1(s_t, Z_t) - L_{accidents}\,.
\label{cost-to-go}
\end{equation} 

To ease the computation burden, we assume that each agent can either drive aggressively or cautiously represented by different values of DRF parameters. For every scenario, DRF controls a total of $M$ agents, which means that there are $2^M$ different combinations of agents' parameters that lead to $2^M$ possible futures. Therefore, the optimal combination corresponding to the most critical traffic scenarios can be obtained with an exhaustive search algorithm. To scalably generate critical scenarios, Simcenter HEEDS \citep{heeds}, a high-performance, global design exploration and optimisation software, is used to optimise the parameters of DRF. The algorithm of generating critical scenarios is shown in Alg. \ref{alg:critical}. 

\begin{algorithm}
\caption{Generate critical scenarios in a model-based multi-agent simulator with DRF}\label{alg:critical}
\begin{algorithmic}[1]
\STATE $\theta \gets \theta_{0}$ // ego vehicle's policy

\FOR{$k=1, ..., N$}
    \STATE // for each traffic scenario
    \STATE // $\Phi^{k}$ are agents' parameters of their DRF policies in the $k^{th}$ scenario

    \STATE // Exhaustively search $2^{M}$ combinations of agents' DRF parameters $\{\Phi_j^k\}_{j=1}^{2^M}$ to get the optimal combination of agents' parameters leading to the critical sceanario
    \STATE $\Phi^k* = \arg \min_{\Phi_j} J(\theta, \Phi_j^k)$
    \STATE // $J$ is computed via \eqref{cost-to-go} by unrolling agents' policies

    \STATE // Validate ego policy $\theta$ in the $k^{th}$ scenario with $M$ DRF agents parameterised by $\Phi^k*$ 
\ENDFOR
\RETURN validation results from critical scenarios
\end{algorithmic}
\end{algorithm}

\subsection{Data augmentation for desired driving behaviours} 
Improving the performance of IL models for AVs is difficult, as adding more training data does not guarantee better performance. To address this problem, here we propose to augment the expert demonstrations by using DRF to control the ego vehicle's velocity profiles along their original trajectories, with different desired driving behaviours encoded in different DRF parameters. This method offers great flexibility to encode desired driving behaviours we wish the IL model to learn. The DRF ensures the new learnt policy is still applicable to the previous dataset because the DRF-augmented data distribution is similar to human demonstrations.

Other data augmentation methods for IL planning models, such as perturbing the original trajectory with noise \citep{Waymo2019}, or requiring an expert policy during training \citep{spl2020}, although they can significantly improve IL performance, they do not guarantee further improvement by retraining with more data. Furthermore, since they cannot be used to learn desired driving behaviours that purposely improve previous weak IL policies, performance usually worsens in critical scenarios where other agents act adversarially. By comparison, our data augmentation method encoding desired driving behaviours can be used to continuously improve poorly trained policies exposed in critical scenarios by learning from DRF-augmented demonstrations. 


\section{Experiments}\label{experiment}

In this section, we evaluate the three contributions of this paper. In particular, we are interested in: the impact of spline trajectory parameterisation on the training efficiency of IL models; the ability of generated critical scenarios to help detect poorly trained policies; and the potential of learning desired driving behaviours via retraining with DRF-augmented demonstrations.











\subsection{Data}
\label{Dataset}
We use the Lyft Prediction Dataset to train and validate our IL models. Both log-replay and generated critical scenarios are used in validation. In log-replay scenarios, the other agents are following their original trajectories. While in critical ones, the other agents are reactive and following the DRF policy, which controls their velocity profiles along their original trajectories.



\subsection{Metrics}
\label{metrics}
We evaluate all models in closed-loop, meaning that the IL policy takes full control throughout the entire duration of each scenario. For each scenario, we measure the following metrics to keep track of the number of violations and events to compare the performance of different IL models.

\begin{enumerate}
\item{\textbf{Safety metrics}: Record the number of \textit{collisions} if the ego vehicle collides with other traffic agents.} 
\item{\textbf{Imitation metrics}: Record the number of \textit{off-road events} if the ego vehicle deviates from its ground-truth trajectory by more than 4m in the lateral direction.}
\item{\textbf{Subjective risk metrics}: Record the number of \textit{aggressive driving behaviours} if the perceived risk (as specified in Sec. \ref{drfmodelling}) of the ego vehicle is larger than $10^{5}$. This is a comprehensive metric that large risk values can mean a very close distance to other vehicles, making the driver feel more at risk.}
\end{enumerate}









\subsection{Data-efficient IL with spline parameterisation}
\label{training efficiency}
In this experiment, we analyse the impact of spline parameterisation on the training efficiency by comparing our data-efficient IL model (trained 30h by us with 1 NVIDIA RTX A4000 laptop GPU, where 1h is equivalent to 10000 iterations, with a batch size of 6 2-second trajectories, i.e., approx. 9 epochs) with Lyft Urban Driver (trained 30h by Lyft with 32 Tesla V100 GPUs, for 61 epochs). To ensure a fair comparison, our IL model shares the same network architecture, except for the dimension of the output of the last layer, as Urban Driver. Both models are trained using the same dataset and simulator from Lyft \citep{Lyft2022}. In Table \ref{tab:imitation learning}, we show that even with significantly less training resources, our model outperforms Urban Driver in all metrics, indicating better performance in safety and imitation. Additionally, our model has a less aggressive driving style compared to Urban Driver.


\begin{table}[]
\tiny
\centering
\caption{Metrics for the baseline and our model from 2500 4-second log-replay scenarios.}
\label{tab:imitation learning}
\resizebox{0.48\textwidth}{!}{
\begin{tabular}{c|ccc|c|c}
\multirow{2}{*}{\textbf{Models}} & \multicolumn{3}{c|}{\textbf{Collision}} & {\textbf{Imitation}} & \multirow{2}{*}{\makecell{\textbf{Aggressive} \\ \textbf{driving}}} \\
             & Front      & Rear & Side       & Off-road    &              \\ \hline
Urban Driver & 1          & 3    & \textbf{0} & 4                & 140          \\ \hline
Ours         & \textbf{0} & 2    & \textbf{0} & \textbf{0}       & \textbf{109} \\ \hline
\end{tabular}
}
\end{table}

\subsection{Generation of critical traffic scenarios}
\label{scenariogeneration}
In this experiment, we compare our generated critical traffic scenarios with adversarial DRF agents to original traffic scenarios with log-replay agents. By unrolling our ego IL policy and Lyft Urban Driver in both kinds of scenarios and comparing their performance, it is shown in Table \ref{tab: data augmentation} that our generated critical scenarios are more challenging for both IL models to handle as the number of collisions increases in critical scenarios.

In addition, more aggressive driving is observed in critical scenarios. This is because the distance between agents is smaller in critical scenarios. Therefore, the ego vehicle subjectively ``feels" more at risk driving in critical scenarios. 


\subsection{Data augmentation for desired driving behaviours}
\label{DA}

Passiveness due to causal confusion is a common mistake of IL models \citep{Causal2019, safetynet2022}, meaning that the IL model may learn to drive passively and unresponsive to approaching vehicles. To mitigate passiveness of the ego vehicle, we retrain the IL model with DRF-augmented data where the ego vehicle drives more aggressively when the rear vehicles are approaching.


In this experiment, we compare the performance of our retrained IL model (30h training + 2h retraining), our IL model (30h training), and Lyft Urban Driver in both log-replay and critical traffic scenarios.


\begin{table}[]
\centering
\caption{Metrics for the baseline and our (retrained) model from 1250 4-second log-replay and critical scenarios.}
\label{tab: data augmentation}
\resizebox{0.5\textwidth}{!}{
\begin{tabular}{l|c|ccc|c|c}
\multirow{2}{*}{\textbf{Scenarios}} &
  \multirow{2}{*}{\textbf{Models}} &
  \multicolumn{3}{c|}{\textbf{Collision}} &
  {\textbf{Imitation}} &
  \multirow{2}{*}{\makecell{\textbf{Aggressive} \\ \textbf{driving}}} \\
                            &              & Front      & Rear       & Side       & Off-road    &             \\ \hline
\multirow{3}{*}{Log-replay} & Urban Driver & \textbf{0} & \textbf{1} & \textbf{0} & \textbf{0}       & 71          \\ \cline{2-7} 
                            & Ours         & \textbf{0} & \textbf{1} & \textbf{0} & \textbf{0}       & \textbf{56} \\ \cline{2-7} 
                            & Ours(Re)     & \textbf{0} & \textbf{1} & \textbf{0} & \textbf{0}       & 100         \\ \hline
\multirow{3}{*}{Critical}   & Urban Driver & \textbf{0} & 8          & \textbf{0} & 6                & 71          \\ \cline{2-7} 
                            & Ours         & \textbf{0} & 7          & 1          & \textbf{0}       & \textbf{60} \\ \cline{2-7} 
                            & Ours(Re)     & \textbf{0} & \textbf{5} & \textbf{0} & \textbf{0}       & 111         \\ \hline
\end{tabular}
}
\end{table}

In Table \ref{tab: data augmentation}, it is shown that our retrained IL model performs better in critical scenarios and equally in log-replay scenarios regarding the collision and imitation metrics. Our retrained model is observed to grasp more aggressive driving behaviours. Also, our retrained model is sufficiently robust to handle both critical and log-replay scenarios. More importantly, we show that driving styles of IL models can be properly customised using DRF without compromising driving safety.


\begin{figure}[]
  \centering
  \includegraphics[width=0.48\textwidth]{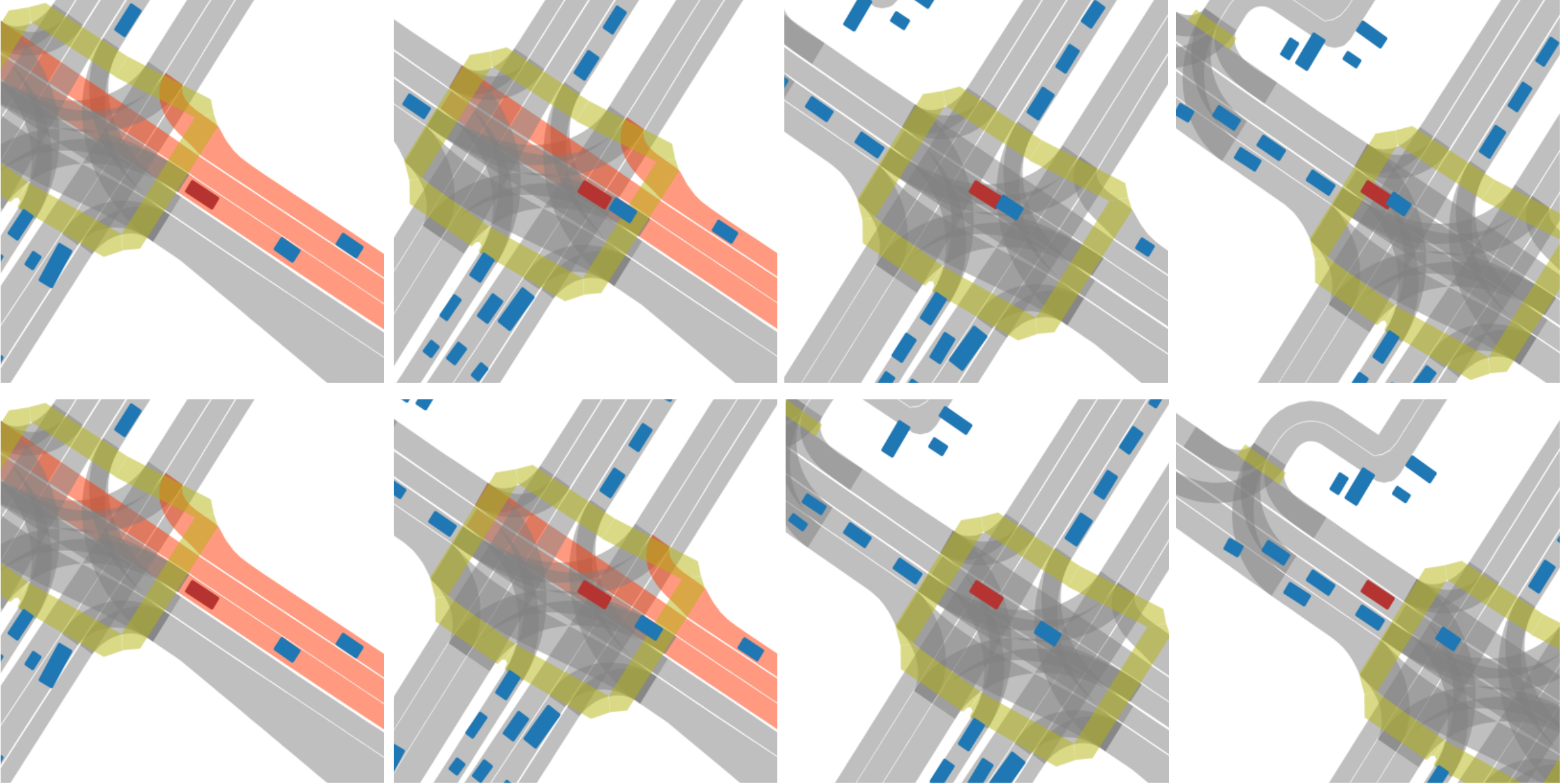}
  \caption{Red rectangle is the ego vehicle controlled by our IL policy, and blue rectangles are other traffic agents controlled by DRFs leading to critical scenarios. Top row: Validation of our IL model in a critical scenario. Bottom row: Validation of our IL model retrained with augmented data in the same critical scenario.}
  \label{retrain}
\end{figure}

Fig. \ref{retrain} presents qualitative results comparing our IL model before and after retraining with DRF-augmented data to alleviate passiveness. In the top row, we see that our IL model before retraining has a rear collision due to passive driving. By contrast, in the bottom row, the retrained model speeds up in time and safely passes the intersection without noticeable sign of passiveness, indicating that it has learnt the desired driving policy that drives slightly more aggressively if followed by a rear vehicle.

\section{Conclusion}

In this paper, we have demonstrated the potential of incorporating the DRF, a parametric human driving behaviour model, in a multi-agent traffic simulator to build a full development cycle that can continuously improve the performance of IL models. With the expressivity and interpretability of the DRF, we can generate critical scenarios with DRF-based agents that are parameterised to act adversarially to the ego IL policy. These generated critical scenarios are proven to be more challenging for the ego IL policy to handle than the recorded scenarios from the logged data. Moreover, weak policies are more easily detected from the validation with critical scenarios. To enhance the weak policy, we use DRF to encode desired driving behaviours to augment the expert demonstrations. By retraining the IL model with augmented data, the IL model achieves safer driving. The IL policy learnt via retraining is more robust as it is applicable to both critical scenarios with adversarial agents and recorded scenarios from logged data. We also show an improved (re)training efficiency by using the spline trajectory parameterisation. For future work, the theory and formulation of DRF should be studied more in-depth to further enhance the realism and diversity of generated critical scenarios.


\bibliography{ifacconf}             
\end{document}